\documentclass[letterpaper, 10 pt, conference]{ieeeconf}  %

\IEEEoverridecommandlockouts                              %

\usepackage{graphics} %
\usepackage[backref=page]{hyperref}
\usepackage{amsmath} %
\usepackage{amssymb}  %
\usepackage{algorithm}
\usepackage[noend]{algpseudocode}
\usepackage[usenames,dvipsnames]{color} 
\usepackage{svg}
\renewcommand{\baselinestretch}{0.95}
\usepackage{subcaption}

\title{\LARGE \bf
R-LGP: A Reachability-guided Logic-geometric Programming Framework for Optimal Task and Motion Planning\\ on Mobile Manipulators
}

\author{Kim Tien Ly$^{1}$, Valeriy Semenov$^{1}$, Mattia Risiglione$^{2}$, Wolfgang Merkt$^{1}$, Ioannis Havoutis$^{1}$%
\thanks{{$^{1}$K. T. Ly, V. Semenov, W. Merkt and I. Havoutis are with the Dynamic Robot Systems (DRS) group, Oxford Robotics Institute, University of Oxford. Email: \tt{\{ktien,valeriy,wolfgang,ioannis\}@robots.ox.ac.uk}.
}}
\thanks{{$^{2}$M. Risiglione is with Istituto Italiano di Tecnologia. Email: \tt{mattia.risiglione@iit.it}.
} }%
}

\begin{document}

\maketitle
\thispagestyle{empty}
\pagestyle{empty}

\begin{abstract}
This paper presents an optimization-based solution to task and motion planning (TAMP) on mobile manipulators. Logic-geometric programming (LGP) has shown promising capabilities for optimally dealing with hybrid TAMP problems that involve abstract and geometric constraints. However, LGP does not scale well to high-dimensional systems (e.g. mobile manipulators) and can suffer from obstacle avoidance issues due to local minima. 
In this work, we extend LGP with a sampling-based reachability graph to enable solving optimal TAMP on high-DoF mobile manipulators. The proposed reachability graph can incorporate environmental information (obstacles) to provide the planner with sufficient geometric constraints. This reachability-aware heuristic efficiently prunes infeasible sequences of actions in the continuous domain, hence, it reduces replanning by securing feasibility at the final full path trajectory optimization.
Our framework proves to be time-efficient in computing optimal and collision-free solutions, while outperforming the current state of the art on metrics of success rate, planning time, path length and number of steps. We validate our framework on the physical Toyota HSR robot and report comparisons on a series of mobile manipulation tasks of increasing difficulty. Videos of the experiments are available \href{https://youtu.be/NEVVHEhQnOQ}{here}.
\end{abstract}

\section{Introduction}
Task and motion planning (TAMP) is the process of decision making that takes a given task as input and outputs a sequence of robot configurations to complete the task. While task planning focuses on high-level task-oriented strategies, motion planning refers to low-level control algorithms that determine the feasibility of robot motion and the continuity considering actuation and joint limits, environmental obstacles, or uncertainties. These methods take into account the variations of sequences of actions that can lead to the desired goal. Therefore, TAMP planners must address both task-level and motion-level requirements, which can be solved either independently or jointly.

A major challenge in TAMP is that high-dimensional geometric constraints in motion control (kinematics, joint limits, reachability, etc.) can limit the effectiveness of high-level strategies. The problem becomes more complex with long-horizon tasks, higher-degrees-of-freedom robots, or multiple--possibly dynamic--obstacles. Therefore, the deliberative function (high-level task planner) should either be informed or have sufficient restrictions on the robot's physical limitations to produce feasible solutions. Consequently, TAMP requires the integration of both high- and low-level planning and TAMP research generally aims to effectively combine both artificial intelligence techniques in task planning and advanced motion planning to tackle such problems. 

In this paper we propose a novel TAMP approach to solve sequential decision-making applications inheriting logic-geometric programming (LGP) \cite{toussaint2015logic}. Our framework generates optimal plans for complex robot tasks, in contrast to current approaches that generate solely feasible solutions, for example, the well-known TAMP solver PDDLStream \cite{garrett2020pddlstream}. Our approach tightly integrates a reachability graph with LGP and allows us to achieve robust, collision-free and kinematically-efficient solutions to LGP on high-dimensional mobile manipulators. We explore the efficacy of our framework in a range of existing TAMP problems of increasing difficulty and report results on success rates, planning times and path lengths. Furthermore, we validated on a physical robot, the Toyota Human Support Robot (HSR), aiming at realistic and practical applications, where geometry and kinematics must be respected. 

\begin{figure}[t]
    \centering
    \begin{subfigure}[b]{0.22\textwidth}
        \centering
        \includegraphics[width=\textwidth]{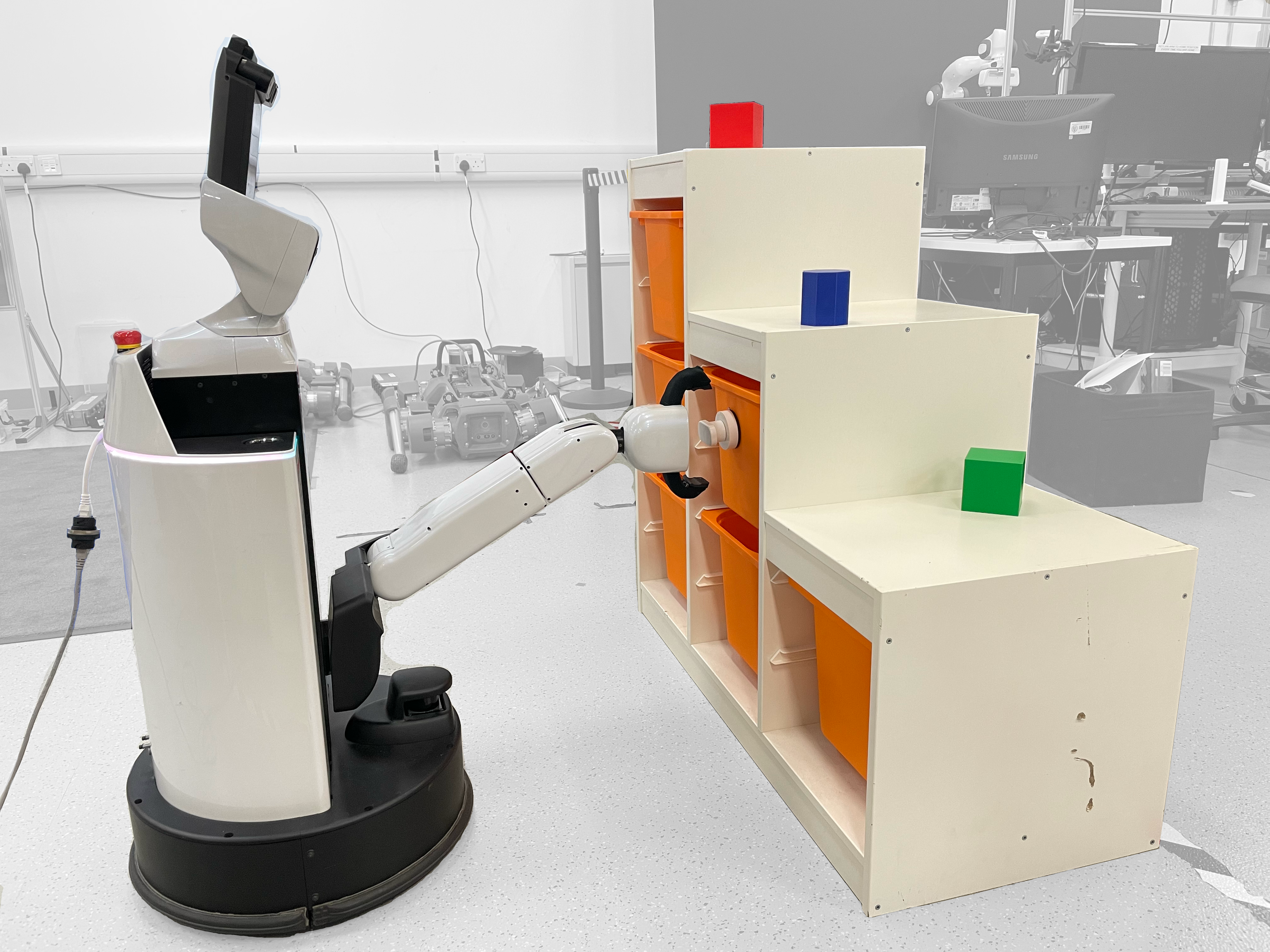}
    \end{subfigure}
    \begin{subfigure}[b]{0.22\textwidth}
        \centering
        \includegraphics[width=\textwidth]{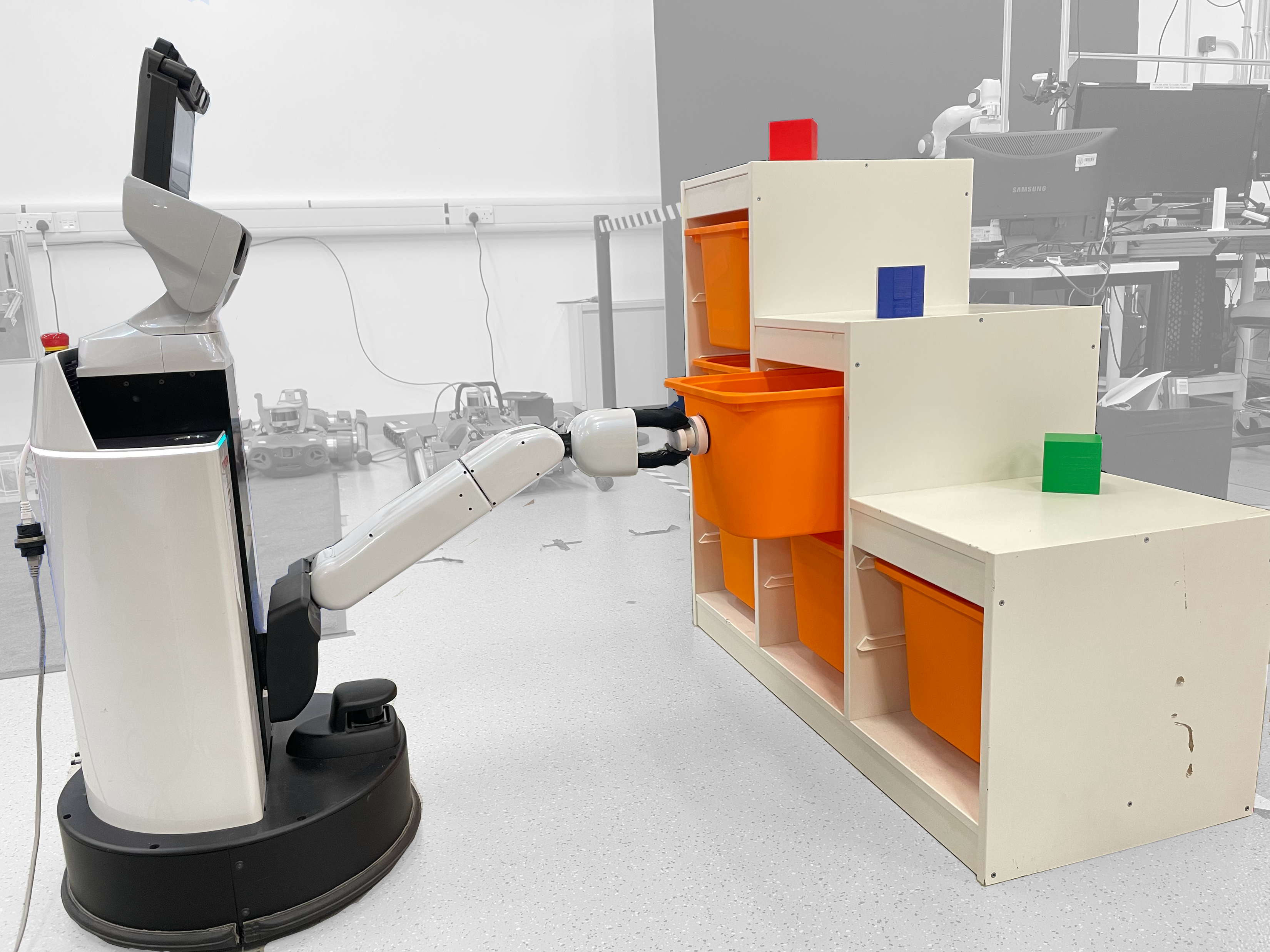}
    \end{subfigure}
    \par\smallskip
    \begin{subfigure}[b]{0.22\textwidth}
        \centering
        \includegraphics[width=\textwidth]{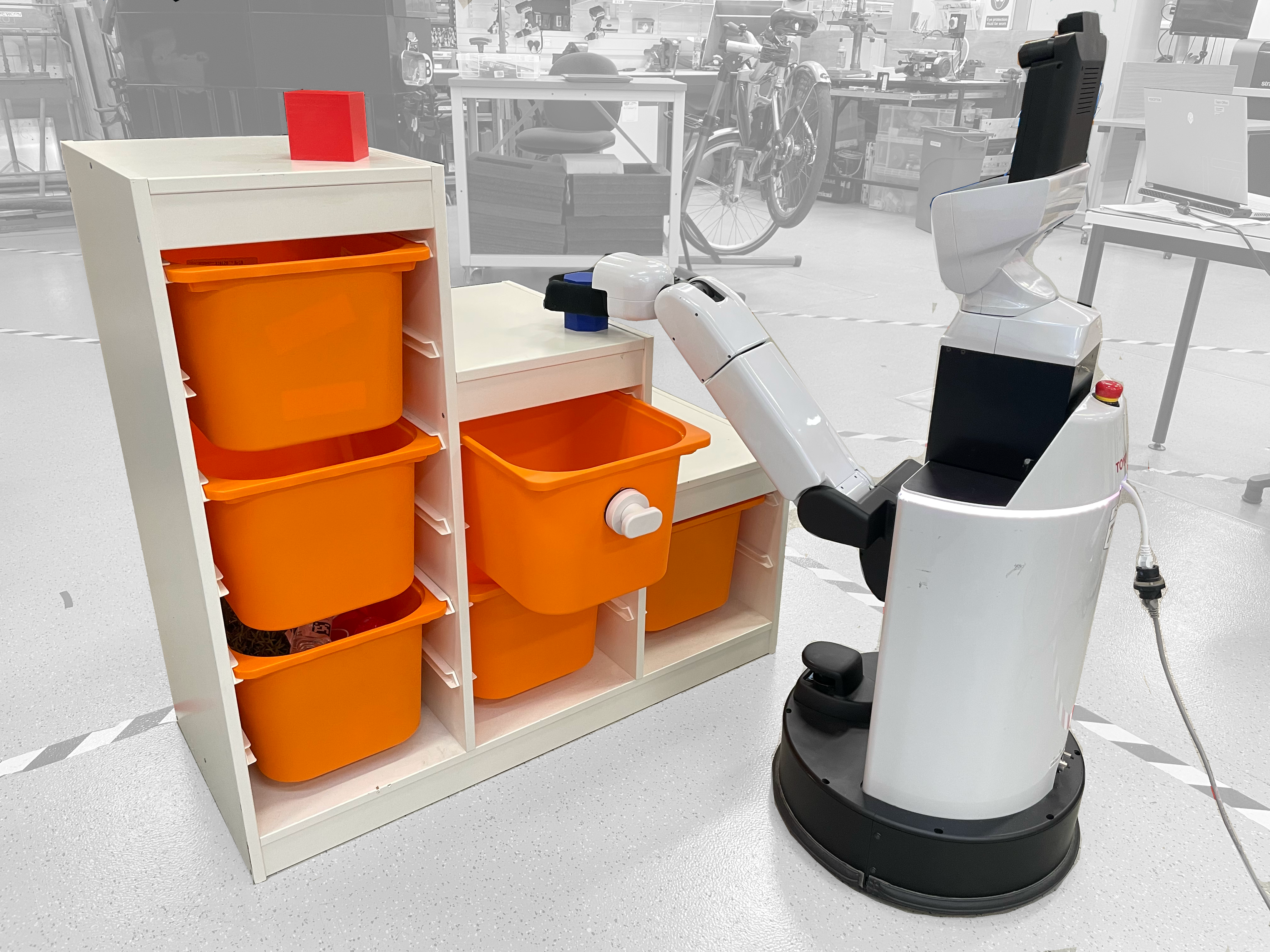}
    \end{subfigure}
    \begin{subfigure}[b]{0.22\textwidth}
        \centering
        \includegraphics[width=\textwidth]{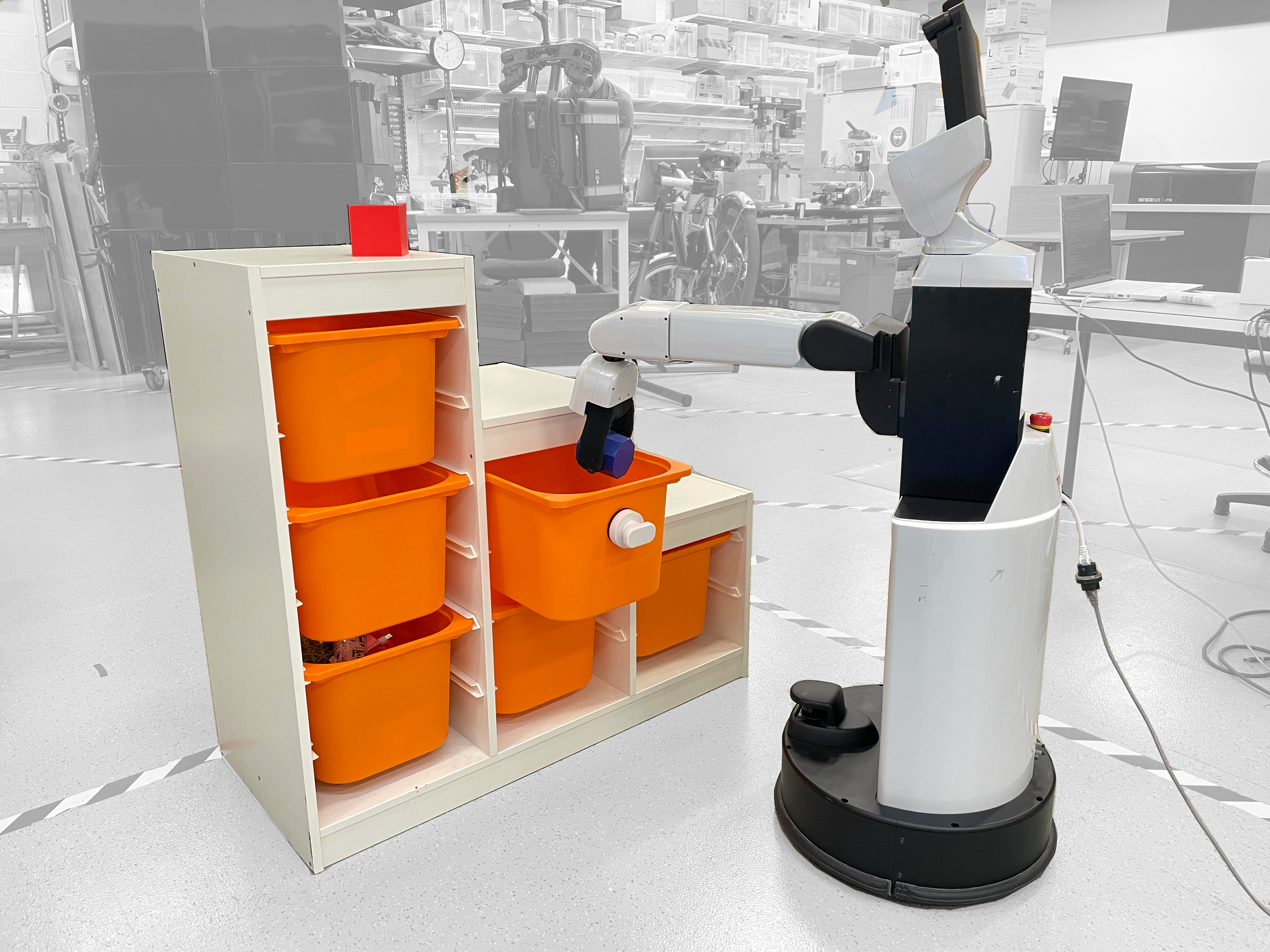}
    \end{subfigure}
    
    \caption{Physical HSR performing R-LGP solution for table clearing task. The captured actions include grabbing knob, opening/closing drawer, picking object, and dropping object.}
    \label{fig:physicalhsr}
\end{figure}

Our main contributions are (1) a LazyPRM-inspired graph for motion planning on floating-base manipulators, (2) a reachability graph serving as symbolic and motion guidance for the LGP planner, (3) a kinematically-effective optimization-based system for TAMP on mobile manipulators. Our novel TAMP approach is a Reachability-guided Logic-geometric Programming framework, in short, \textit{R-LGP}.

\section{Related work}

\subsection{Satisfactory TAMP}
The main challenge in combining task and motion planning is their hybrid nature. While task planning involves discrete task specifications, motion planning is typically solved in continuous space.  Task and motion planning are traditionally combined on a level basis, which can be decoupled or integrated. Accordingly, each layer will keep its own level of abstraction and the TAMP model should be able to translate knowledge between the two levels. Given the introduction of STRIPS (Stanford Research Institute Problem Solver) \cite{fikes1971strips}, robot planning was initially studied with a decoupled hierarchical structure \cite{Nils1984}. 
The approach in this early work assumes that motion control can solve all high-level actions without alternative backtrack solutions.
Meanwhile, integrated TAMP \cite{Garrett2021} considers failures and can replan to ensure that the strategy is executable. Common methods are taking one of the two parts as the base plan and solving the other level accordingly. For example, Srivastava et al. \cite{Srivastava2014} introduces a TAMP method that backtracks and finds alternative task plans if the motion solver fails. Such approach usually discretizes the continuous space for symbolic planning. The method can utilize off-the-shelf planners and state-of-the-art techniques. TMKit \cite{dantam2018tmkit} is introduced as the first open-source framework to implement such a structure, based on \cite{dantam2016incremental}\cite{dantam2018incremental}. There is an opposite approach where the task search space is bounded by sampling the continuous workspace \cite{Thierry2004}.  Recently, Kim et al. \cite{kim2023reachability} introduced a reachability tree-based sampling algorithm that pre-generates goal state to bias action search. Given that task planning is usually in a finite domain, having to re-plan the task sequences is considered to produce a lower cost compared to continuously checking geometrical feasibility. 

PDDLStream \cite{garrett2020pddlstream} extends Planning Domain Definition Language (PDDL)\cite{mcdermott1998pddl} by using streams to enable sampling procedures. This work proves to be a modular and domain-independent approach to TAMP. 

\subsection{Optimal TAMP}
To reduce complexity, TAMP approaches usually tackle task and motion planning separately and focus more on completeness. Solving such problems optimally is difficult because the optima of each layer might not be the overall optimal solution. An example can be found in \cite{kim2023}, where the authors used a two-layer hierarchical structure of optimization to find asymptotically optimal solutions to coverage planning task. Kongming \cite{li2008generative} was the first approach to integrate continuous control variables in an activity planner. The method combines a Planning Graph for discrete actions and Flow Tubes for continuous actions, which was modeled as a mixed logic linear (non-linear) program. The method ties to a fixed time discretization, which is difficult to apply on long-horizon tasks. However, it brings up the need for a tightly coupled task and motion planning framework in robotics applications. 

Toussaint \cite{toussaint2015logic} introduced the logic-geometric programming (LGP) algorithm, which formulates TAMP as a mathematical program that uses optimization methods to find locally optimal plans instead of solely feasible ones. The task in this work is to build the highest stable tower from a list of blocks and boards.  By bringing logic into geometry, LGP does not need to arbitrarily discretize the continuous space and directly optimize continuous solutions. An extended work from the team \cite{hartmann2020robust} integrated RRT to reconstruct a large pavilion. Although the method solves the issue when the optimizer fails in a cluttered environment, the system ignores system uncertainties. Variants of LGP papers include a heuristics method for long-horizon tasks \cite{braun2022rhh}\cite{driess2019hierarchical}, a dynamic LGP for human motion prediction \cite{le2021hierarchical} or an approximation solver for cooperative manipulation \cite{toussaint2017multi}. Other optimization-based mathematical approaches to TAMP is constraint programming (CP) \cite{behrens2019constraint} \cite{booth2016mixed} and mixed integer programming (MIP) \cite{conforti2014integer} \cite{booth2016mixed} \cite{ioan2021mixed}. CP is a more general term than MIP, Booth et al. \cite{booth2016mixed} did a comparison between MIP and CP and concluded that inference-based CP is better than relaxation-based MIP in terms of time and solution quality. This work is tested on simulation and claims to be case-sensitive. MIP is originally a decision making and scheduling method that solves non-convex problems. When it comes to robotics planning, this optimization solution can help to capture discrete decisions with integer variables. Deits et al. \cite{deits2014footstep} introduced a planner that uses MIP to generate globally optimal sequences of footsteps in difficult terrain. Multi-robot task planning is a common application of MIP \cite{culbertson2019multi}\cite{lippi2021mixed}. In short, these mathematical techniques allow encoding logical decisions and geometric constraints in nonlinear optimization models without backtracking, targeting globally optimal strategies. 

As more interest has been driven to solving TAMP optimally, besides mathematical programming, a recent work from Thomason et al. \cite{thomason2022task} proposes asymptotically optimal decision-making using informed tree search. The model effectively combines constraint-based symbolic planning, distance-based predicate representation, and batch-sampling-based optimal motion planning to solve a hybrid state space problem. Recently, inspired by LGP, Sleiman et al. \cite{sleiman2023versatile} introduced an offline bilevel optimization planner that solves multi-contact problems on loco-manipulation system. The system successfully plans whole-body motion for a quadrupedal mobile manipulator to open/close doors and dishwashers.

\section{Proposed framework overview}

In LGP, the authors solve TAMP with three levels of approximation and switches between symbolic search and configuration optimization that is conditioned by symbolic decisions. While \textit{level 1} decides the symbolic sequence, \textit{levels 2} and \textit{3} optimize keyframes (e.g. grasp poses) and full path respectively. An extended work, RHH-LGP \cite{braun2022rhh}, proposes a heuristic to guide the symbolic search in level 1. The kinematic reachability here, which is an important factor to the feasibility of the action sequences, is fixed or calculated using Euclidean distance or the length of robot's links. 
In this work, we extend LGP with a sampling-based reachability graph to enable solving optimal TAMP on high degrees-of-freedom (DoF) mobile manipulators. The proposed reachability graph can also incorporate environmental information (obstacles) to provide the planner with sufficient geometric constraints. This reachability-aware heuristic efficiently prunes infeasible sequences of actions in the continuous domain. Hence, it reduces replanning by securing feasibility at the final full trajectory optimization.

\begin{algorithm}[tb]
\caption{Reachability-guided LGP}\label{alg:reachability}
\begin{algorithmic}[1]

\Require symbolic goal $g$, initial symbolic state $s_0$, initial configuration $x_0$, world $W$
\State $s \leftarrow s_0 $
\While{not $s \in S_{goal}(g)$}
    \State PathFound $\leftarrow$ False
    \State SymbolicSearch $\leftarrow s$ 
    \State Switch $\leftarrow$ ${\o}$    
    \State Path $\leftarrow$ ${\o}$
    \While {not PathFound}
        \State \textit{// Construct reachability graph}
        \State \textbf{RG} = constructRG($W$)
        \State \textit{// Symbolic search}
        
        \State $n$ $\leftarrow$ SymbolicSearch.argmin(heuristicCost(\textbf{RG}))
        \State Switch.append($n$)
        \If{not $n$ $\in S_{goal}(g)$}
            \State SymbolicSearch.append($n$.expand())
        \Else
        \State \textit{// Optimize over kinematic switches}
        \If{SwitchOptimization(Switch)}
            \State waypoints $\leftarrow$ getWaypoints(\textbf{RG}, $n$)
            \State Path.append(waypoints)
        \State \textit{// Optimize over the full path}
        \If{PathOptimization(Path)}
            \State PathFound $\leftarrow$ True
            \State $s \leftarrow n$
        \EndIf
        \Else
            \State break
        \EndIf
        \EndIf
\EndWhile
\EndWhile
\end{algorithmic}
\end{algorithm}

Our reachability-guided LGP (R-LGP) pipeline is described in Algorithm \ref{alg:reachability}. The proposed reachability graph serves as a guide to the first (symbolic) and final (full path) layers in LGP. In the pipeline, nodes chosen from the symbolic search, with the help of our reachability heuristic, will be sent to the kinematic switch optimization. Once a feasible sequence is found, the system will solve the full path optimization with the guided path from the precomputed reachability graph. The main contribution of the graph is twofold: \textit{a)} to provide a heuristic that informs the LGP's symbolic planner about kinematics and geometry; \textit{b)} to provide collision-free guidance to the LGP's trajectory optimization at level 3 (full path), in order to avoid local optima.

\section{Reachability Graph}
The reachability graph is designed in a LazyPRM-like \cite{bohlin2000path} manner. We compute the graph with node validation and edge checking during planning.
\subsection{Graph generation}
This section explains the \textbf{constructRG} function in Algorithm \ref{alg:reachability}.
\subsubsection{Node sampling}
Nodes are randomized in task space
$x \in \mathbb{R}^{3}$ with uniform distribution $U(p_{\text{min}}, p_{\text{max}})$. Nodes in the graph represent end-effector positions of the mobile manipulator, which sufficiently denotes robot reachability. We validate nodes with a constrained optimization formulation. The robot model is defined as a loco-manipulation system with a floating-base torso as in Eq. \ref{eq:q}, where $m$ is the number of DoFs of the manipulator. This allows the framework to be implemented on either legged or wheeled mobile platforms.  In our work, the mobile platform is holonomic and does not have any restrictions on translation or rotation, however, these can be added in as constraints. 
\begin{equation}
  \begin{aligned} 
    \label{eq:q}
    & q=[q_{\text{base}},q_{\text{manipulator}}], \\ 
    \text{where } & q_{\text{base}} = [x_{\text{base}},y_{\text{base}},\psi_{\text{base}}],\\
    & q_{\text{manipulator}} = [q_1, q_2, q_3, ..., q_m]. 
    \end{aligned}
\end{equation}

The nonlinear optimization formulation for node validation is defined in Eq. \ref{eq:ik}, where $x_0$ is the checked node, representing the reference for the robot end-effector position. $g_{\text{prec}}()$ checks collision between the inspecting node and the environment, and determines the precondition for the optimization with a constant $M\rightarrow +\infty$. This condition removes unreachable nodes for the sake of processing time. On the other hand,  $f_{\text{kin}}()$ computes the end-effector position $x$ that corresponds to the joint values $q$. Function $g()$ defines inequality constraints for whole-body joint limits and collision. In the reachability graph, orientation of the end-effector is not constrained for flexibility and generality. Our graph is designed to generate a guiding cost and path for LGP. The full trajectory optimization, with orientation constraints, will then be computed on the final layer with constraints for the relevant manipulation action.
\begin{equation}
    \begin{aligned} 
        \label{eq:ik}
        \min_{q}\; & M\cdot\max(0, g_{\text{prec}}(x_0)) + \left| x - x_0 \right|^2\\
    \text{s.t.  }& g_{\text{prec}}(x_0)\le 0,\\
                & x=f_{\text{kin}}(q), \\
               & g(q) \le 0. 
    \end{aligned}
\end{equation}
After collision and inverse kinematics (IK) validation is performed, the node is added to the graph's node list $N$ as $n(x, q, c)$. While $x\in \mathbb{R}^{3}$ is the end-effector of the robot, $q\in \mathbb{R}^{3+m}$ denotes the optimized IK configuration. $c$ equals to the cost value from the optimization solution. The sampling process stops when the number of $N$ reaches a predefined number $N_{\text{node}}$.

\subsubsection{Edge connections}
As mentioned, we relax edge connection by assuming all edges are collision-free. For each sampled node in the list $N$, we connect to $k$ nearest neighbours without checking for collision. Each edge is added to the graph with the cost defined in Eq. \ref{eq:edgecost}. The cost includes a combination of weighted Cartesian and configuration Euclidean distances between two end nodes, along with nodes' costs. $w_x$, $w_q$ and $w_c$ are the correspondent weights for the Cartesian, configuration distance and IK cost. 
\begin{equation}
    \begin{aligned} 
      c_e = w_x \cdot \left \| x_1 - x_2  \right \| + w_q \cdot\left \| q_1 - q_2 \right \| + w_c \cdot (c_1 + c_2)
    \label{eq:edgecost}
    \end{aligned}
\end{equation}

\subsection{Path querying}
\subsubsection{Solution library}
The reachability graph runs as an underlying service, waiting for the LGP planner to query  two ends of a path. The service leverages a library to store previously computed solutions to avoid replanning for the same path and save planning time. During symbolic search in the first LGP's layer, the planner might iterate through different logical sequences and query a path in opposite directions. For instance, a first call may seek a path from A to B, and subsequently, a solution from B to A. This is beneficial for tasks at the same levels, where one does not have a precondition on another action's completion.\\
The structure of each solution is shown in (\ref{eq:datastructure}), where \textit{key} contains the start and end point of the path, \textit{path} and \textit{cost} come from the resulting solution with $n$ being the number of nodes on the path.  Noted that the final configuration in \textit{path} $q_{1n}$ matches $x_2$ Cartesian pose. Each \textit{key} is unique across the library in sorted order, and is used to query the data bidirectionally. On each call, the service checks the library and returns the stored solution. If the requested \textit{key} is new to the library, the system starts to query the graph for the path. The answer is then added to the library for future reference.
\begin{equation}
    \begin{aligned} 
        \label{eq:datastructure}
        \text{solution = }\{ %
        & \text{key: } (x_1, x_2) \;, x_i \in \mathbb{R}^3\\
        & \text{path: } [q_{10}, q_{11}, ..., q_{1n}] \in \mathbb{R}^{n \times(3+m)}\\
        & \text{cost: } h%
        \}
    \end{aligned}
\end{equation}
 In this data structure, \textit{cost} is returned if the pipeline is querying heuristic cost, which is the function \textbf{heuristicCost} in Algorithm \ref{alg:reachability}. When the framework reaches the final layer, \textit{path} serves as guidance for trajectory optimization, accessed through the function \textbf{getWaypoints}.

\subsubsection{Path planning}
The reachability graph planner receives two Cartesian positions for each path querying message \textit{key} as starting point and ending point, naming  $x_k$ and $x_{k+1}$. Nodes corresponding to $x_k$ and $x_{k+1}$ are added to the graph upon solving (\ref{eq:ik}). Collision is checked for connecting these two nodes to the existing graph.
If either $x_k$ or $x_{k+1}$ is unable to connect to the graph, we start enhancing the node. Gaussian sampling is applied with in-loop decreased covariance to find adjacent nodes as waypoints to connect to the existing graph. During this process, collision checking on both nodes and edges is performed to ensure a meaningful enhancement. 

We use Dijkstra for finding the shortest path in the graph. We verify the path by checking collisions on all connected edges. We linearly interpolate the trajectory in configuration space with a predefined number of steps $L$. An edge with a detected collision is removed from the graph and replanning is triggered. In the case that any node loses all connections due to edge removal, it will be enhanced with the same method as starting/ending node enhancement. The difference is that the inspecting node can be replaced with an effective sampled one, which has the lowest IK cost $c$ and is able to connect to the graph. In this sense, we ensure that the graph is fully connected, meaning that no subset is disconnected from the rest of the graph. During planning, the system updates the graph with enhancement and edge removal, benefiting future planning queries.

\begin{figure}[tb]
    \centering
    \includegraphics[width=0.48\textwidth,trim={1.25cm 0cm 0cm 0cm},clip]{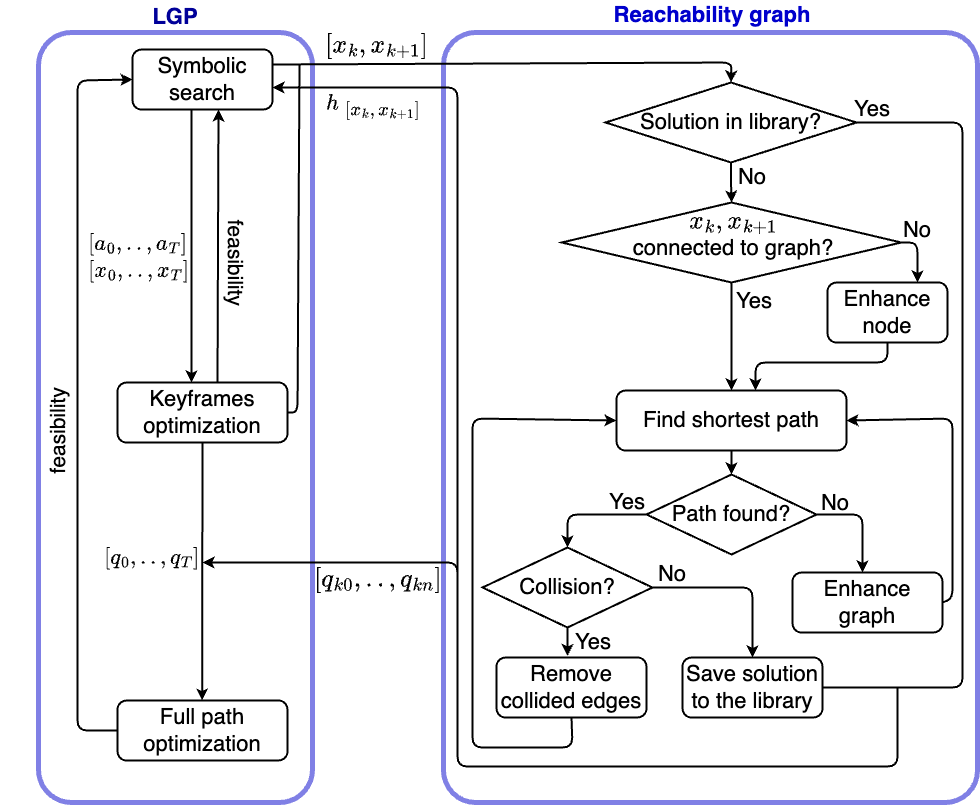}
    \caption{Path querying system in R-LGP.}
    \label{fig:path_planning_chart}
\end{figure}

The flowchart in Fig. \ref{fig:path_planning_chart} shows in detail the path planning algorithm along with the interactions between the reachability graph and the LGP framework. The heuristic cost $h$ is used internally in the symbolic search level. Before entering into the last stage of LGP, we interpolate $[q_{k},q_{k+1}]$ ($k \in [0, T-1]$) with corresponding $[q_{k0},..,q_{kn}]$ as input guidance for full path optimization.

\section{Evaluation}
\subsection{Simulation evaluation}
In order to validate our approach, we propose a set of simulations and hardware experiments, performed on the PR2 and HSR robots. Our baseline for comparison is the heuristic-based LGP - RHH-LGP \cite{braun2022rhh}. In addition, we compare our system against PDDLStream \cite{garrett2020pddlstream}, the most popular and ready-to-use TAMP solver. We implemented the adaptive version of PDDLStream, which claims to outperform all other approaches for cost-sensitive problems. This allows us to set a priority on minimizing path length. There is an \textit{optimal} flag in PDDLStream solver which determines whether or not the planner should explore more options and conclude with the best solution. We name \textit{PDDLStream nonoptimal} and \textit{PDDLStream optimal} for PDDLStream with \textit{optimal} set to false and true respectively. 

The experiments are designed to address the ability in accommodating geometric and kinematic constraints in integrated TAMP problems. We conducted an extensive evaluation with 100 runs in simulation for each experiment. All random tasks are intentionally feasible. The recorded metrics are:
\begin{itemize}
    \item \textit{Success rate}: the number of collision-free TAMP solutions over the total number of runs.
    \item \textit{Planning time}: average time taken to get to successful solutions.
    \item \textit{Path length}: average length, in meters, of the robot's base trajectory, assuming a floating-base robot.
    \item \textit{Number of steps}: average number of steps to complete the given task. For this metric, we only consider key switches, e.g. pick, place, which intuitively correspond to the number of objects.
\end{itemize}

\subsubsection{Pick and place}
In this scenario, the PR2 robot is asked to pick and place 3 objects on the tables to the tray. This is a default TAMP task in the PDDLStream framework and is widely used in TAMP development \cite{garrett2020pddlstream}. The objects are at random positions on the table over 100 runs. We set a fixed table size and all random poses are reachable from one side of the table. 

\begin{figure}[tb]
    \centering
    \begin{subfigure}[b]{0.402\textwidth}
        \centering
        \includegraphics[width=\textwidth]{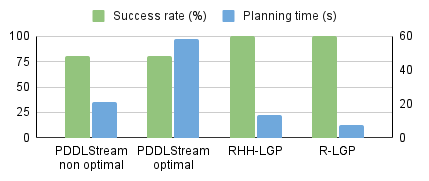}
    \end{subfigure}
    \begin{subfigure}[b]{0.402\textwidth}
        \centering
        \includegraphics[width=\textwidth]{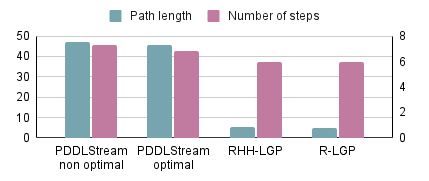}
    \end{subfigure}
    
    \caption{Comparison result of the TAMP planners on pick and place task.}
    \label{fig:picknplaceresult}
\end{figure}

The results in Fig. \ref{fig:picknplaceresult}  show that the PDDLStream optimal solution is better in path length and step count than the default PDDLStream, although the latter achieves a shorter planning time. With PDDLStream solutions, PR2 tends to travel to the side of the table nearest to the object to pick it up regardless of reachability. LGP frameworks provide shorter trajectories by not navigating around the table, hence, also reduce the execution time. Apart from optimality, as opposed to PDDLSream, LGP frameworks achieve 100\% success rate. Our R-LGP framework effectively reduces planning time leveraging the reachability heuristic. 

\subsubsection{Sorting}
We increase the complexity of both task planning and motion planning in this task. The robot is asked to sort objects into trays with the same colors. Unlike the classic sorting task, we extend the experiment to address the reachability and mobility awareness of the system. In particular, we use larger table tops in this test, and randomized objects can be out-of-reach from one side of the table. This task requires TAMP solvers to accommodate obstacle-free configuration trajectories in solving high-level tasks, specifically requiring the robot to navigate around the table.  To ensure all tasks are feasible, we constraint the objects' randomized locations to be within reachable distance from at least one side of the table.

\begin{figure}[h]
    \centering
    \begin{subfigure}[b]{0.402\textwidth}
        \centering
        \includegraphics[width=\textwidth]{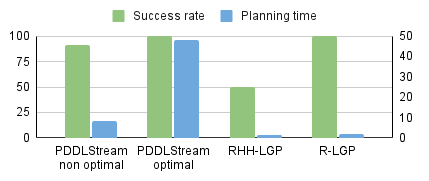}
    \end{subfigure}
    \begin{subfigure}[b]{0.402\textwidth}
        \centering
        \includegraphics[width=\textwidth]{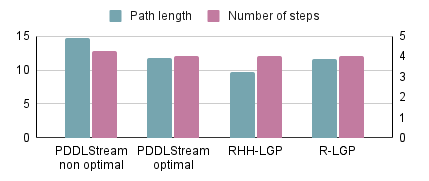}
    \end{subfigure}
    
    \caption{Comparison result of the TAMP planners on sorting task.}
    \label{fig:sortingresult}
\end{figure}

In Fig. \ref{fig:sortingresult}, it is worth noticing that the baseline RHH-LGP fails to plan a collision-free trajectory in 50\% of the cases. Due to local optima in trajectory optimization, RHH-LGP planner can only solve tasks where the randomized location is within reach from the same side to the current position. This issue then filters runs with shorter path and results in a shortest average path length. PDDLStream optimal and R-LGP provide similar quality of final solutions with the same success rate of 100\%. However, the planning time in R-LGP is significantly reduced from 48s to 2s. 

\begin{figure}[tb]
    \centering
        \begin{subfigure}[b]{0.18\textwidth}
        \includegraphics[trim={150 0 150 60},clip,width=\textwidth]{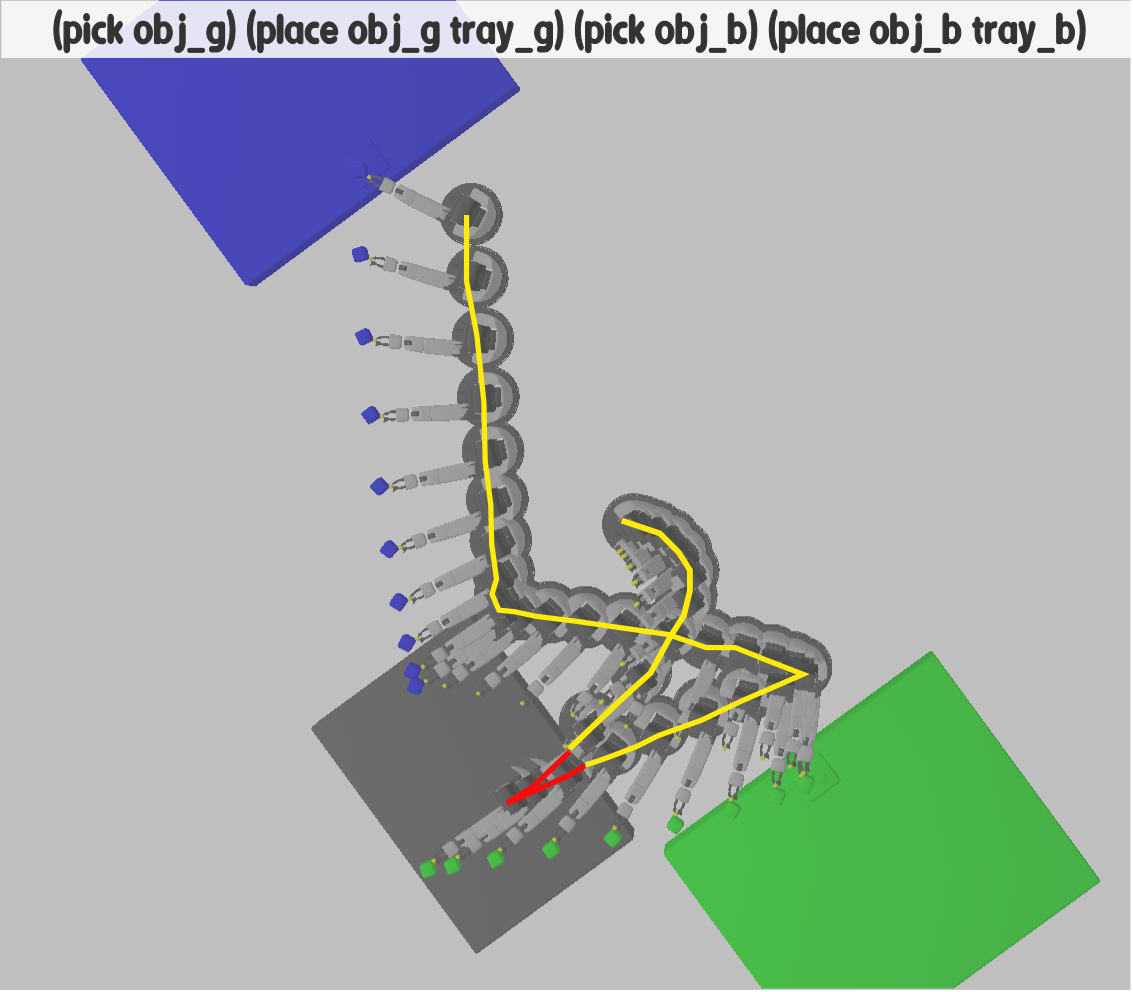}
        \caption{RHH-LGP}
        \label{fig:rhhlgp_path}
    \end{subfigure}
    \begin{subfigure}[b]{0.18\textwidth}
        \includegraphics[trim={150 0 150 60},clip,width=\textwidth]{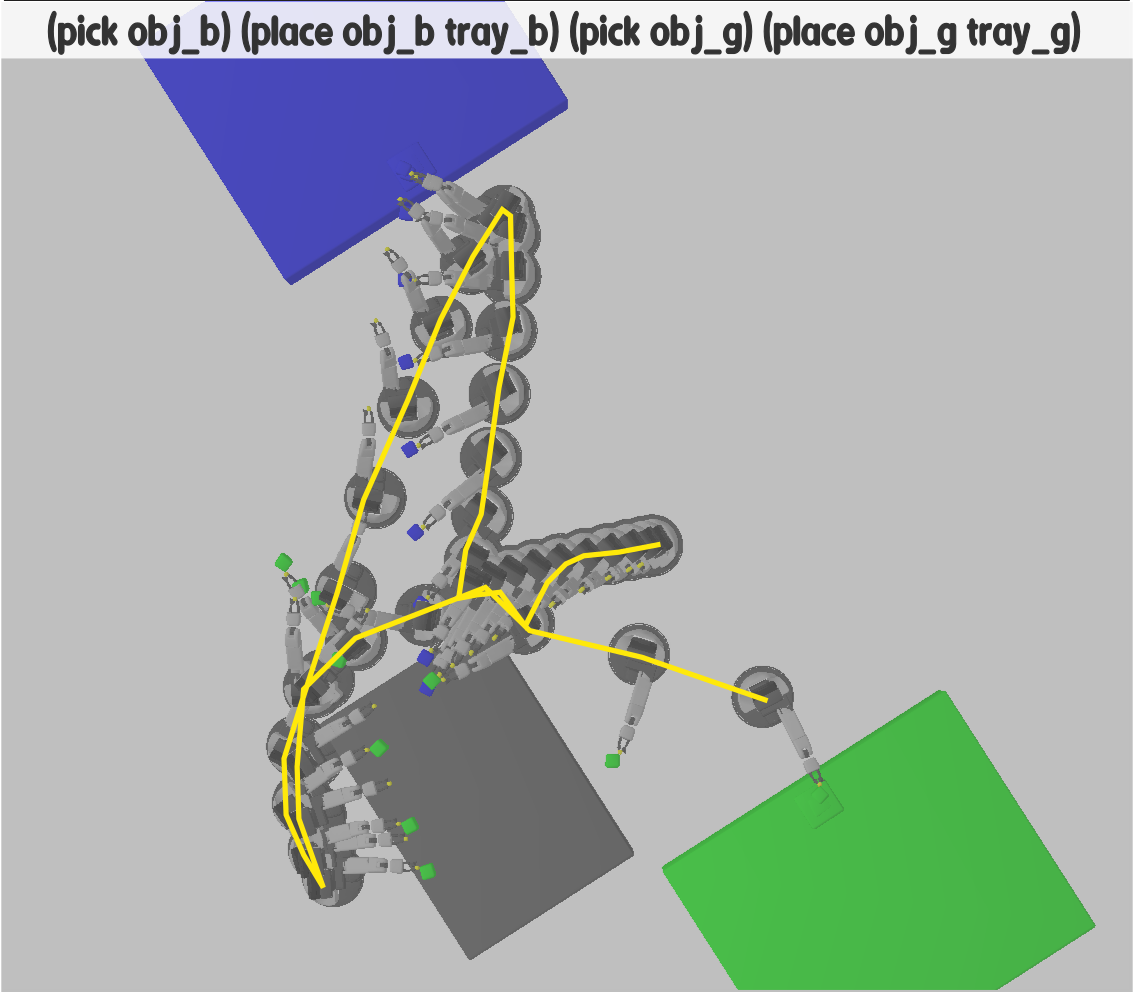}
        \caption{R-LGP}
        \label{fig:rlgp_path}
    \end{subfigure}
    \caption{TAMP results. Objects are initially spawned on the grey table and should be put on the table with the same color. R-LGP outperforms RHH-LGP in generating reachability-aware robot base trajectory (yellow lines) and avoiding collisions (red lines).}
    \label{fig:path}
\end{figure}

Fig. \ref{fig:path} captures an example of TAMP results for the sorting task with the LGP baseline (a) and our planner (b). In the figures, the robot's base trajectory is shown in yellow line, which also denotes the resulted task sequence. In RHH-LGP, the planner considers the distance-based heuristic, hence, decides to sort the green block first. Meanwhile, our heuristic cost informs the high-level planner that the robot must go around the table in order to pick up the green block. Therefore, the decision to pick the blue block first produces a shorter travelling cost. With the help of the reachability graph, the symbolic planner can consider the cost of navigating taking into account reachability capabilities. In terms of full trajectory optimization, the red line shown in Fig. \ref{fig:rhhlgp_path} is the violation of collision checking in RHH-LGP. We successfully solve this issue with the guidance from the reachability graph to the final layer in LGP, which enables the collision-free path in Fig. \ref{fig:rlgp_path}. 

\subsection{Real robot validation}
\begin{figure}[tb]
    \centering
        \begin{subfigure}[b]{0.16\textwidth}
        \includegraphics[trim={0cm 4cm 0cm 16cm},clip,width=\textwidth]{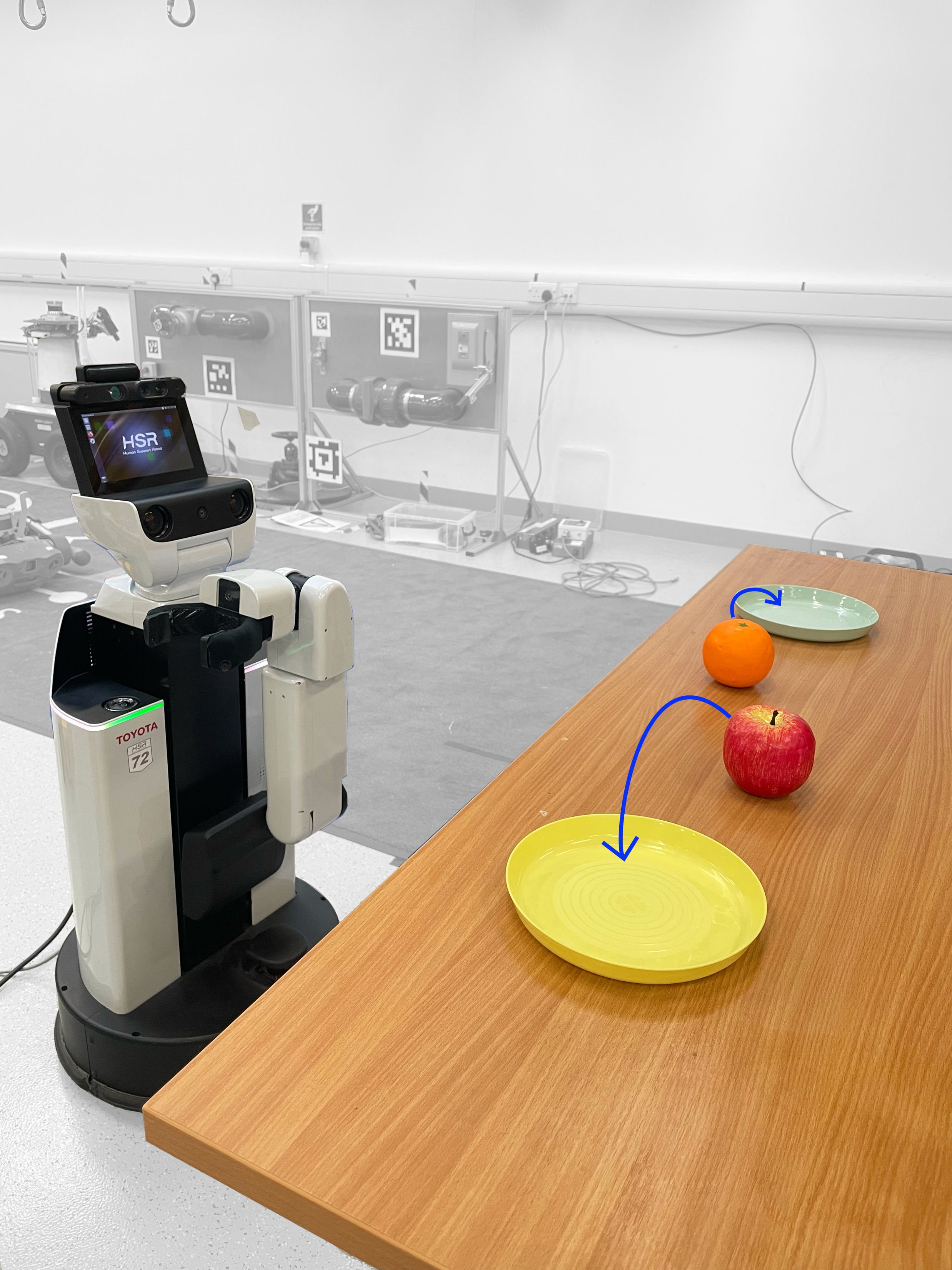}
    \end{subfigure}
    \begin{subfigure}[b]{0.16\textwidth}
        \includegraphics[trim={0cm 4cm 0cm 16cm},clip,width=\textwidth]{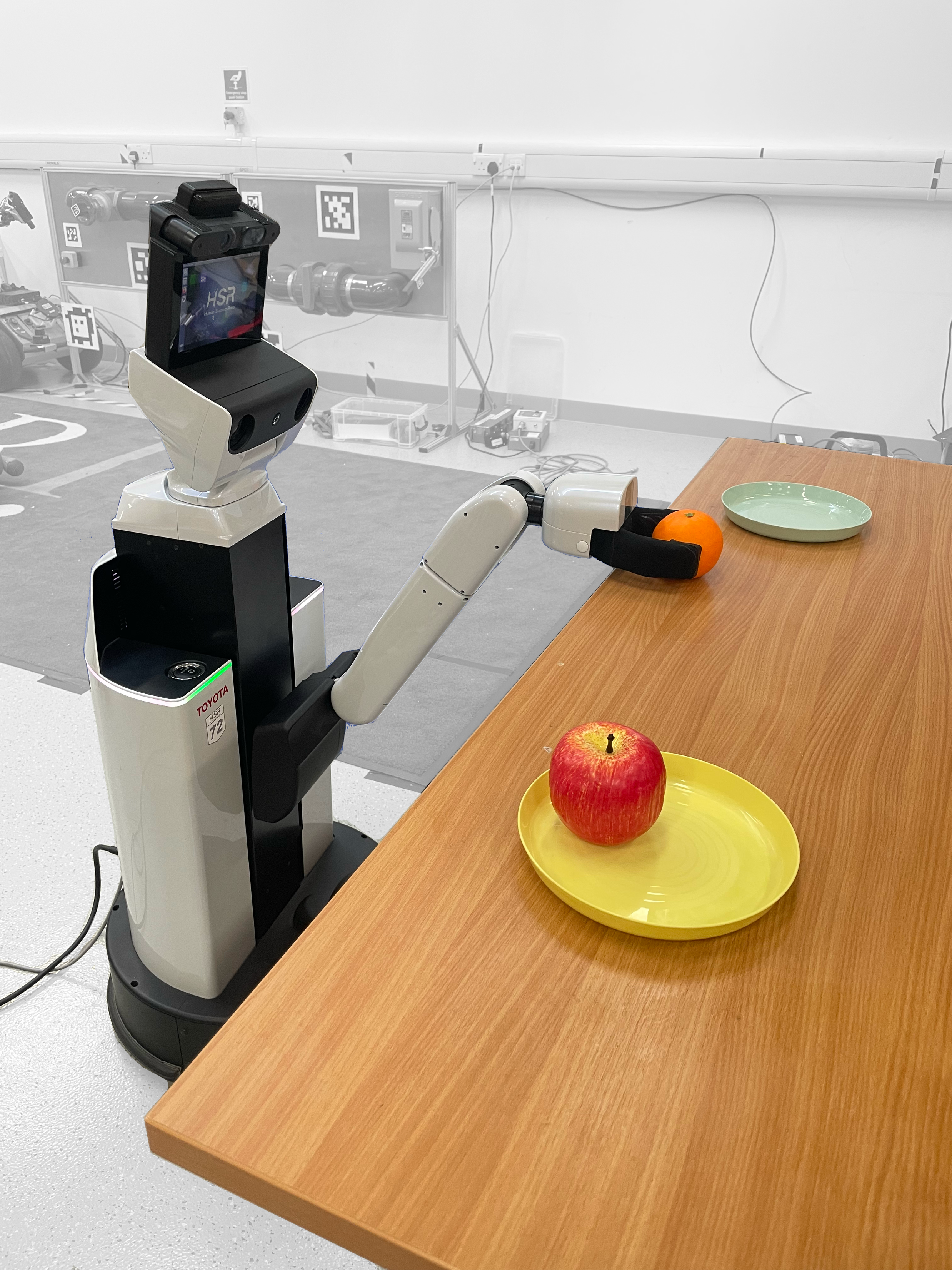}
    \end{subfigure}
    \caption{Physical HSR using R-LGP solver in sorting task.}
    \label{fig:physicalhsr_sorting}
\end{figure}

We validated our framework on a physical HSR. We implemented 2 tasks of increasing difficulty:
\begin{itemize}
    \item \textit{Sorting task}: the robot is asked to pick objects and place in the correct trays (Fig. \ref{fig:physicalhsr_sorting}).
    \item \textit{Table clearing}: the task is to put objects from the surface into the drawer (Fig. \ref{fig:physicalhsr}). 
\end{itemize}

In this real-world experiment, R-LGP proved to successfully solve a longer-horizon task with table clearing. The desired symbolic sequence of this mission is (\textit{take\_knob, open\_drawer, pick\_object, drop\_object, take\_knob, close\_drawer}). Figure \ref{fig:physicalhsr} provides a visualization of the listed behaviours. Treating \textit{take\_knob} separate to \textit{open\_drawer} and \textit{close\_drawer}, the work can also be applied to opening/closing drawers, cabinets or doors. In our trial, the stair-like cabinet, as shown in Fig. \ref{fig:physicalhsr}, also poses a challenge of reachability awareness and obstacle avoidance. When the object is placed on the lower step, picking up action must consider that certain directions are blocked by the step above it.

In the sorting task, the R-LGP planner generated the result in 2s for 4 steps, with 2 objects correctly placed. It took longer for table clearing, 17s, to plan the long sequence with 6 separate steps, which the robot travelled 9m. In both examples, with our R-LGP solutions, the HSR robot can perform the task robustly with an optimal and collision-free trajectory. 
\subsection{Reachability graph evaluation}

For each environment, we precompute the reachability graph only once. On average, the reachability graph is built within 14.2s with 200 samples over the specified workspaces.

\begin{figure}[tb]
    \centering
        \begin{subfigure}[b]{0.35\textwidth}
        \centering
        \includegraphics[width=\textwidth,trim={7cm 7cm 7cm 5cm},clip]{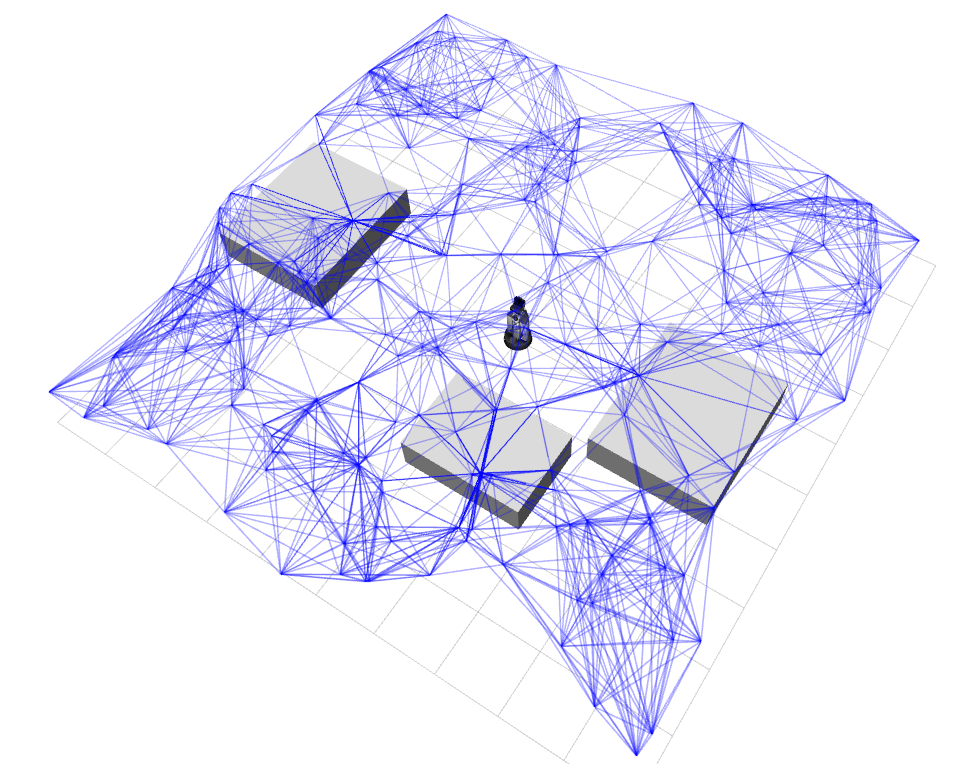}
        \caption{Pre-built}
        \label{fig:rgraph_prebuilt}
    \end{subfigure}
    \hfill    
    \begin{subfigure}[b]{0.35\textwidth}
        \centering        \includegraphics[width=\textwidth,trim={7.5cm 8cm 8cm 4.5cm},clip]{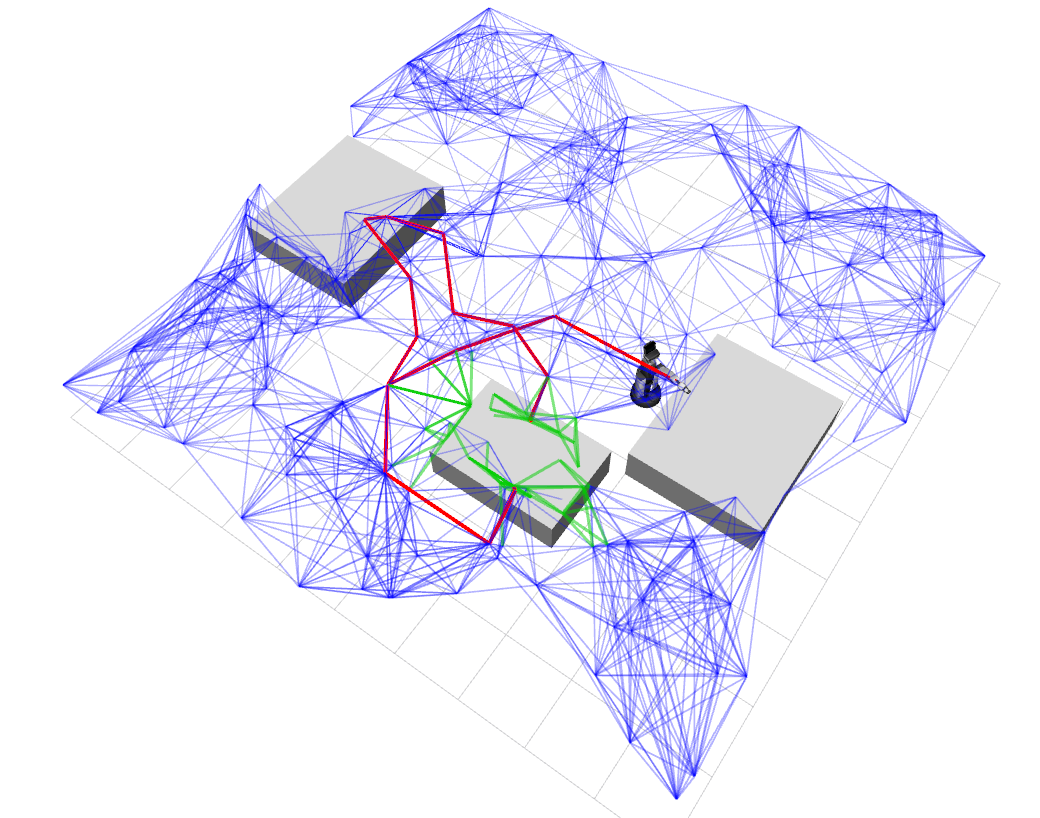}
        \caption{After planning}
        \label{fig:rgraph_planned}
    \end{subfigure}    
    \caption{Reachability graph before and after planning. The red line is an example of the generated guided path, and green lines present the graph enhancement over time.}
    \label{fig:rgraph}
\end{figure}

We designed the sorting task to evaluate the robustness and completeness of our reachability graph, as this environment has more geometric constraints to consider during motion planning. Figure \ref{fig:rgraph} visualizes our reachability graph for this scenario. The pre-built reachability graph is shown in Fig. \ref{fig:rgraph_prebuilt}. In this step, nodes are validated with a collision-free configuration state and neighbour edges are all connected. The generated solution for the whole successful trajectory is plotted in the red line in Fig. \ref{fig:rgraph_planned}. The solution sufficiently produces a collision-free guidance path that has an understanding of the robot kinematics and reachability. This resulting path provides LGP with meaningful heuristic cost and trajectory guidance. The enhanced graph after a few queries is also denoted in Fig. \ref{fig:rgraph_planned}. The green lines represent added nodes and edges while searching for paths. In this example, the enhancement is mainly around the middle table, as objects for sorting are randomized on this table.
The graph is updated over time, where edges in collision are removed and new nodes are added. This reduces planning time for subsequent queries in the same workspace.

\section{Conclusion}
R-LGP, our proposed framework, provides a robust solution to TAMP on mobile manipulators, which is tightly integrated to LGP. The introduction of the reachability graph not only successfully informs the LGP symbolic planner with information regarding reachability and mobility capabilities but also resolves local optima in trajectory optimization. The reachability-aware heuristic cost helps to prune out infeasible trajectories at the first layer, hence, reducing failures at the final full trajectory optimization. Despite being a sampling-based graph, it boosts the efficiency of both symbolic search and trajectory optimization without loss of optimality for LGP. As a result, the system is time-efficient in generating optimal TAMP solution while respecting robot geometry and kinematics. 
We presented an extensive evaluation to validate the framework, including real-world experiments on the Toyota HSR robot. In comparison to the state-of-the-art PDDLStream in the basic pick and place task, we reduced path length by 10 times with 6 times quicker solving time. The sorting task experiment highlighted the drawbacks of the current LGP framework, where environmental constraints can adversely affect symbolic planning. Without guidance, the baseline also fails to generate a collision-free trajectory. Hence, we achieved twice better success rate and improved completeness of the LGP solver. While sampling-based motion planning does not provide optimal solutions in limited time, our proposed reachability graph integrated to LGP can solve optimal TAMP.  An avenue for future work is pruning task-specific constraints for TAMP on long-horizon applications. In our current framework, we are addressing the motion rather than task knowledge, and we believe pruning will help with scaling up to larger state spaces.

\newpage
\bibliographystyle{IEEEtran}
\bibliography{IEEEabrv, references}
\end{document}